\documentclass[letterpaper]{article} % DO NOT CHANGE THIS
\usepackage{aaai25}  % DO NOT CHANGE THIS
\usepackage{times}  % DO NOT CHANGE THIS
\usepackage{helvet}  % DO NOT CHANGE THIS
\usepackage{courier}  % DO NOT CHANGE THIS
\usepackage[hyphens]{url}  % DO NOT CHANGE THIS
\usepackage{graphicx} % DO NOT CHANGE THIS
\usepackage{natbib}  % DO NOT CHANGE THIS AND DO NOT ADD ANY OPTIONS TO IT
\usepackage{caption} % DO NOT CHANGE THIS AND DO NOT ADD ANY OPTIONS TO IT
\usepackage{amsthm}
\newtheorem{theorem}{Theorem}
\usepackage{graphicx}
\usepackage{amsmath}
\usepackage{times}  % DO NOT CHANGE THIS
\usepackage{helvet}  % DO NOT CHANGE THI
\usepackage{courier}  % DO NOT CHANGE THIS
\usepackage[hyphens]{url}  % DO NOT CHANGE THIS
\usepackage{graphicx} % DO NOT CHANGE THIS
\usepackage{natbib}  % DO NOT CHANGE THIS AND DO NOT ADD ANY OPTIONS TO IT
\usepackage{caption} % DO NOT CHANGE THIS AND DO NOT ADD ANY OPTIONS TO IT
\usepackage{algorithm}
\usepackage{algorithmic}
\usepackage{soul}
\usepackage{url}
\usepackage[utf8]{inputenc}
\usepackage{graphicx}
\usepackage{amsmath}
\usepackage{amsthm}
\usepackage{booktabs}
\usepackage{algorithm}
\usepackage{algorithmic}
\usepackage[switch]{lineno}

\usepackage{graphicx}
\usepackage{amsmath}
\usepackage{amssymb}
\usepackage{booktabs}
\usepackage{dsfont}
\usepackage{amsfonts,amssymb}
\usepackage{graphics}
\usepackage{tikz}
\usepackage{multirow, graphicx}
\usepackage{colortbl}
\usepackage{bbding}
\usepackage{bm}
\usepackage{times}
\usepackage{soul}
\usepackage{url}
\usepackage[utf8]{inputenc}
\usepackage{amsmath}  % For better math support
\usepackage{graphicx} % For \scalebox
\usepackage{amsmath}

\newtheorem{proposition}{Proposition}
\newtheorem{hypothesis}{Hypothesis}
\usepackage[switch]{lineno}
\usepackage{algorithm}
\usepackage{algorithmic}
\usepackage{booktabs}
\usepackage{amsmath} 
\usepackage{amssymb}
\usepackage{amsfonts}
\usepackage{times}  % DO NOT CHANGE THIS
\usepackage{helvet}  % DO NOT CHANGE THIS
\usepackage{courier}  % DO NOT CHANGE THIS
\usepackage[hyphens]{url}  % DO NOT CHANGE THIS
\usepackage{graphicx} % DO NOT CHANGE THIS
\usepackage{natbib}  % DO NOT CHANGE THIS AND DO NOT ADD ANY OPTIONS TO IT
\usepackage{caption} % DO NOT CHANGE THIS AND DO NOT ADD ANY OPTIONS TO IT
\usepackage{textcomp}
\usepackage{graphicx}
\usepackage{amsmath}
\usepackage{amsfonts} 
\usepackage{multirow, graphicx}
\usepackage{colortbl}
\usepackage{bbding}
\usepackage{tabularx}
\usepackage{graphicx}
\usepackage{amsmath}
\usepackage{booktabs}
\usepackage{dsfont}
\usepackage{amsfonts,amssymb}
\usepackage{graphics}
\usepackage{tikz}
\usepackage{ragged2e}

\usepackage{algorithm}
\usepackage{algorithmic}

\usepackage{newfloat}
\usepackage{listings}
\usepackage{multirow}
\usepackage{amssymb}  % 提供了数学符号，包括 \Checkmark
\usepackage{array}      % 高级表格
\usepackage{newfloat}
\usepackage{listings}
\DeclareCaptionStyle{ruled}{labelfont=normalfont,labelsep=colon,strut=off} % DO NOT CHANGE THIS
\floatstyle{ruled}
\newfloat{listing}{tb}{lst}{}
\floatname{listing}{Listing}
\title{Toward Modality Gap: Vision Prototype Learning for Weakly-supervised Semantic Segmentation with CLIP}
\author{
    Zhongxing Xu\textsuperscript{\rm 1}\equalcontrib,
    Feilong Tang\textsuperscript{\rm 1}\equalcontrib,
    Zhe Chen\textsuperscript{\rm 2},
    Yingxue Su\textsuperscript{\rm 2}, \\
    Zhiyi Zhao\textsuperscript{\rm 3}, 
    Ge Zhang\textsuperscript{\rm 2},
    Jionglong Su\textsuperscript{\rm 2},
    Zongyuan Ge\textsuperscript{\rm 1}\thanks{Corresponding author: Zongyuan Ge}
}
\affiliations{
    \textsuperscript{\rm 1}AIM Lab, Faculty of IT, Monash University\\
    \textsuperscript{\rm 2}Xi’an Jiaotong-Liverpool University\\
    \textsuperscript{\rm 3}University of Edinburgh\\
    Zhongxing.Xu@monash.edu
}
\usepackage{bibentry}
\begin{document}

\maketitle

\begin{abstract}
The application of Contrastive Language-Image Pre-training (CLIP) in Weakly Supervised Semantic Segmentation (WSSS) research powerful cross-modal semantic understanding capabilities. Existing methods attempt to optimize input text prompts for improved alignment of images and text, by finely adjusting text prototypes to facilitate semantic matching. Nevertheless, given the modality gap between text and vision spaces, the text prototypes employed by these methods have not effectively established a close correspondence with pixel-level vision features. In this work, our theoretical analysis indicates that the inherent modality gap results in misalignment of text and region features, and that this gap cannot be sufficiently reduced by minimizing contrast loss in CLIP. To mitigate the impact of the modality gap, we propose a Vision Prototype Learning (VPL) framework, by introducing more representative vision prototypes. The core of this framework is to learn class-specific vision prototypes in vision space with the help of text prototypes, for capturing high-quality localization maps. Moreover, we propose a regional semantic contrast module that contrasts regions embedding with corresponding prototypes, leading to more comprehensive and robust feature learning. Experimental results show that our proposed framework achieves state-of-the-art performance on two benchmark datasets.\end{abstract}


\begin{figure}[t]
\centering
\begin{tabular}{cc}
\includegraphics[width=0.46\textwidth]{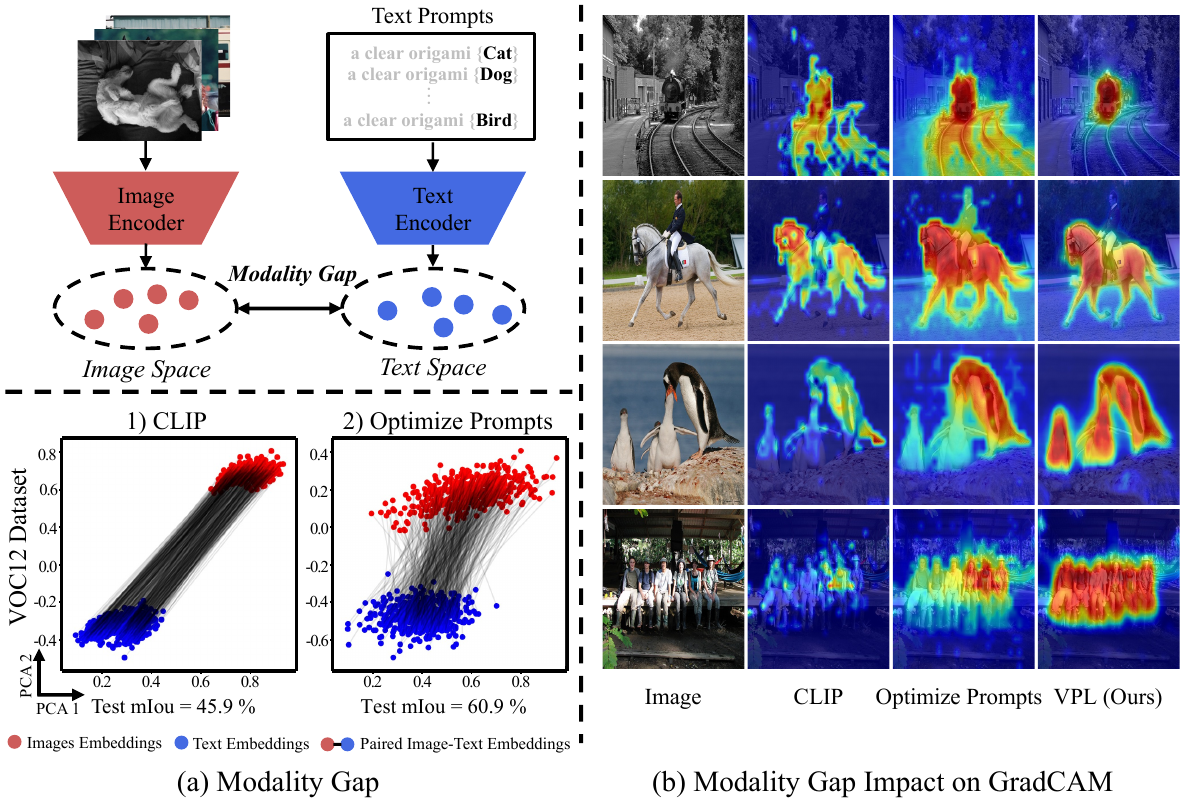}
\end{tabular}
\caption{The main idea proposed in this paper is to reduce the impact of the modality gap by learning vision prototypes. We show the modality gap between paired image embeddings and text prototypes. Even though the Optimized prompt minimizes contrastive loss between prototypes and object region, the modality gap is still not sufficiently reduced (as shown in (a)), which results in the text prototypes failing to accurately capture the relevant region of the target object. In contrast, we propose VPL, which ensures accurate activation of the complete object region (b).
}

\label{intro}
\end{figure}

\section{Introduction}
\quad Semantic segmentation serves as a fundamental task in the field of computer vision~\cite{tang2023duat,wang2022stepwise,xu2024polyp,hu2025ophnet,hu2024ophclip,yang2023action,xiong2024sam2, trinh2024sight,tang2024discriminating,hu2024diffusion,zhang2024magic,wan2024sigma}. Weakly Supervised Semantic Segmentation (WSSS) has gained popularity in the research community. It learns from weak labels such as image-level labels~\cite{lee2021anti}, scribbles~\cite{lin2016scribblesup}, or bounding boxes~\cite{lee2021bbam}, instead of pixel-level annotations. Most WSSS approaches utilize CAM~\cite{zhou2016learning} to provide localization cues for target objects, thereby mapping visual concepts to pixel regions. After training with 400 million image-text pairs, Contrastive Language-Image Pre-training~\cite{radford2021learning} has established a strong association between vision concepts and text descriptions, demonstrating excellent cross-modal semantic matching capabilities. Inspired by transfer learning, several WSSS methods ~\cite{yang2024foundation} apply powerful text prototypes encoded by the CLIP model to generate high-quality CAMs.

The key to WSSS is to generate CAM that effectively covers the complete object. Recent weakly supervised research ~\cite{lin2023clip,xu2023learning,liu2023referring,yang2024foundation} primarily focus on constructing more adaptive text prototypes by optimizing text cues, with the aim to establish stronger correlations with semantic object regions for dense localization. Customized text prompting strategies can enhance the semantic alignment between object regions and their corresponding text prototypes to a certain extent, but a significant semantic gap still persists. This gap is due to the inherent cone effect of CLIP, which causes the modality gap between text and vision spaces ~\cite{liang2022mind}. As shown in Fig~\ref{intro} (a), even after minimizing the contrastive loss between the text prototype and the semantic object region, the cone effect remains significant.

The semantic misalignment caused by the modality gap affects the discriminative ability of the text prototypes in two aspects. On the one hand, our theoretical analysis indicates that text prototypes contain inactive and redundant information. Since CLIP was trained with millions of captions and concepts, its text embeddings contain a large number of features that may not be relevant to a specific target domain ~\cite{hu2024reclip}. However, the categories in the target domain usually require the activation of only a limited set of specific category features, resulting in a large amount of unnecessary text information in the text prototypes. On the other hand, text prototypes lack vision details such as lighting, color, and texture. This absence weakens their ability to match the detailed visual features, leading to a deficiency category representativeness of text prototypes. As shown in Fig~\ref{intro} (b), although the optimized text prototypes show effectiveness, they still struggle to capture more complete and accurate target object areas. Moreover, they erroneously activate similar or highly co-occurring categories (\textit{e.g.,} \texttt{railroad} and \texttt{train} in Fig.~\ref{intro} (b)). Therefore, the text prototypes are affected by the modality gap leading to suboptimal segmentation performance of the model.

To address these challenges, we propose a \textbf{V}ision \textbf{P}rototype \textbf{L}earning (VPL) framework to reduce the impact of the modality gap by introducing more representative vision prototypes. We theoretically demonstrate that the optimal category prototypes for vision tasks should originate from the vision space. Moreover, our analysis reveals that the modality gap between text and vision spaces obtained by CLIP is inherent, confirming that optimal vision prototypes cannot be achieved in text space. Our VPL framework is comprised of two main phases. In the first phase, instead of taking the conventional approach of directly using text prototypes, we use text prototypes with KL-divergence as a constraint to precisely learn vision prototypes in the vision space through gradient descent. These vision prototypes are then utilized to regenerate more accurate pseudo-masks, as shown in Fig.~\ref{intro} (b). In the second phase, we use the refined pseudo-masks to effectively supervise the decoder, transforming high-level feature information into accurate pixel-level predictions. However, imperfect mask labels provide ambiguous knowledge to the segmentation network. To address the issues, we propose regional semantic contrast that contrasts corresponding mask region embedding and vision prototypes, driving the network to learn more comprehensive and robust object embeddings and enhancing adaptability to the modality gap.

% has two main technical novelties.
In the PASCAL VOC 2012 \cite{everingham2010pascal} and MS COCO 2014 \cite{lin2014microsoft} datasets, we evaluate our method in various WSSS settings, where our approach achieves state-of-the-art performance. The key contributions of our research can be summarized as follows:
\begin{itemize}
\item We are the first to investigate the impact of modality gap in weakly-supervised semantic segmentation with CLIP and introduce the Vision Prototype Learning framework to alleviate this issue.
\item We theoretically demonstrate the inherent modality gap in CLIP and propose to obtain vision prototypes from the vision space with help of the text prototypes.
\item We introduce a regional semantic contrast module to enhance the alignment between object regions and vision prototypes. Experiments demonstrate the efficacy of our framework with state-of-the-art performance on mainstream benchmarks.
\end{itemize}

\section{Related Work}

\noindent \textbf{Weakly Supervised Semantic Segmentation} as an important topic in computer vision\cite{ye2024CSSL,chen2024CTTA,zhang1,zhang2,zhang3,zhang4}, can generally be divided into single-stage and multi-stage learning processes. Single-stage methods \cite{wang2023treating,wu2024masked} use image-level labels to train the segmentation network in an end-to-end manner. In contrast, multi-stage methods \cite{rong2023boundary,wu2023hierarchical,zhou2022regional,cho2024finding} first generate segmentation seeds and then use them as pseudo-labels to train off-the-shelf segmentation network for better performance. In order to generate high-quality seeds, recent research has proposed solutions such as adversarial erasure \cite{yoon2022adversarial}, region growing \cite{peng2023usage}, localizing attention map \cite{xu2024mctformer+}, and exploring boundary constraints \cite{ru2023token}.  
\\ \hspace*{\fill} \\
\noindent \textbf{Contrastive Language-Image Pre-training} achieves strong generalization ability through training with a large number of vision-text pairs. Although CLIP excels in cross-modal learning,~\cite{liang2022mind} indicates that there is an inherent gap between different data modalities within the CLIP model, leading to insufficient alignment between text and visual representations. Modality gap have been intensively studied in fields such as few-shot learning \cite{ouali2023black}, zero-shot learning \cite{qian2023intra}, and domain adaptation \cite{hu2024reclip}. To mitigate this issue, SuS-X \cite{udandarao2023sus} introduces additional images to provide richer visual information. In recent developments, CLIP has been widely applied in WSSS. CLIMS ~\cite{xie2022clims} introduces a loss function derived from CLIP to guide a network to produce high-quality CAM. CLIP-ES ~\cite{lin2023clip} has developed GradCAM and Class-Aware Attention Affinity (CAA) techniques for directly creating CAM using CLIP. Meanwhile, WeCLIP ~\cite{zhang2024frozen} leverages a frozen CLIP model to extract semantic features and combines it with a new decoder and refinement module (RFM) to generate and correct pseudo labels. In addition to these approaches, FMA ~\cite{yang2024foundation} proposes two sets of task-specific learnable text prompts to capture category semantic knowledge relevant to segmentation, thereby constructing more representative text category prototypes.

\section{Preliminary}
Given each training image $x_i \in \mathbb{R}^{w \times h \times 3}$ in the dataset $\left\{x_i\right\}_{i=1}^I$, where $w$ and $h$ are the width and height of the images, respectively, and $I$ is the total number of images. Each image is associated with only an image-level label $\left\{z_n\right\}_{n=1}^N$ in which $N$ is the pre-specified categories, and their corresponding vision and text representations can be extracted as:
\begin{equation}\small
\mathbf{x}_i=E_v\left(x_i\right) ; \quad \mathbf{z}_n=E_t\left(z_n\right); \quad M_n= \text{GradCAM}(\mathbf{x}_i,\mathbf{x}_n), \label{extracted}
\end{equation}
where $E_v(\cdot)$ and $E_t(\cdot)$ denote the image and text encoder in the pre-trained CLIP model, respectively. $\mathbf{x}_i \in \mathbb{R}^{\text{S} \times \text{H} \times D}$ and $\mathbf{z}_n \in \mathbb{R}^{D}$ represent the vision features and the text features with $D$ channels. $\mathbf{x}_i$ has a spatial size of $\text{S} \times \text{H}$, and $M_n$ denotes a GradCAM. See Appendix A for more details.

\noindent \textbf{Modality Gap in CLIP.}\label{3.2} Given a batch of $B$ $\left(v_i,t_i\right)$ pairs, CLIP learns to maximize the cosine similarity of the image and text embeddings of the $B$ real pairs in the batch while minimizing the cosine similarity of the embeddings of the $B^2-B$ incorrect pairs. The model learns two encoders by minimizing the contrastive loss $\ell$, which can be written as:

\begin{equation}\small
\begin{aligned}
\ell\left(\mathbf{v}_i, \mathbf{t}_i\right) = & \sum_i - \log \frac{\exp \left(\mathbf{v}_i^{\top} \mathbf{t}_i / \tau\right)}{\sum_j \exp \left(\mathbf{v}_i^{\top} \mathbf{t}_j / \tau\right)} \\
& - \log \frac{\exp \left(\mathbf{t}_i^{\top} \mathbf{v}_i / \tau\right)}{\sum_j \exp \left(\mathbf{t}_i^{\top} \mathbf{v}_j / \tau\right)}.
\end{aligned}
\end{equation}
where $\mathbf{v}_i$ and $\mathbf{t}_i$ represent the L2-normalized embeddings of the image and text in the $i$-th pair, respectively. $\tau$ is a learned temperature parameter to scale the logits. The cross-entropy loss compares the similarity between image and text embeddings, focusing on cross-modal association and ranking. 

A significant difference between inter-modal and intra-modal distributions in CLIP has been observed~\cite{liang2022mind,udandarao2023sus}. The Euclidean distance between modalities, $\emph{i.e.,}$ $\sum_i\left\|\mathbf{v}_i-\mathbf{t}_i\right\|_2^2$, is related to the magnitude of temperature. However, the small $\tau$ does not pull the text and vision space together (the proof in Appendix B.1).
The CLIP model focuses on high-level associations between images and texts rather than fine-grained vision features, and it does not merge images and texts into a unified, indistinguishable representation space, as shown in Fig.~\ref{intro} (b). This inability to map to the visual features required for precise segmentation means that the prototypes from the text space may not capture the nuances of the vision space, resulting in degraded performance for segmentation tasks.

\section{Methodology}

\noindent \subsection{Learning Vision Prototypes from Text Supervision}\label{3.3}
To mitigate the impact of the modality gap, we propose to introduce more representative prototypes to better capture the spatial coverage of semantic object regions. 
We utilize mask labels $\{m_{i}\}$ for the standard semantic segmentation supervised learning in vision space as follows:
\begin{equation}\small
\ell\left(\mathbf{x}_i, m_{i,k} \right) = \min _W \sum_{i,k} -\log \left(\frac{\exp \left({\mathbf{x}_{i,k}}^{\top} \mathbf{w}_{m_{i,k}} / \tau_I\right)}{\sum_n \exp \left({\mathbf{x}_{i,k}}^{\top} \mathbf{w}_n / \tau_I\right)}\right),
\label{33}
\end{equation}
where $\mathbf{w}_n$ denotes the learnable vision classifier, and $W=\{\mathbf{w}_n\}^N_{n=1} \in \mathbb{R}^{d \times N}$, $\mathbf{x}_{i,k}$ denotes the $k$-th pixel representation in image $i$. $\tau_I$ is the temperature parameter for the optimization with vision data. Minimizing the objective function in Eq.~\eqref{33} requires the adjustment of $\mathbf{w}_n$. This causes the \textbf{decision boundary} of $\mathbf{w}_n$ to generate higher scores in the feature space $\{\mathbf{x}_i\}$, with the aim to maximally reflect the semantic information and category distribution of the images.

% where $\mathbf{w}n$ is the learnable vision classifier and $W = {\mathbf{w}n}{n=1}^N \in \mathbb{R}^{d \times N}$. $\mathbf{x}{i,k}$ denotes the $k$-th pixel representation in image $i$, and $\tau_I$ is the temperature parameter for vision data optimization. Minimizing the objective function in Eq.~\eqref{33} adjusts $\mathbf{w}_n$ to improve the decision boundary, aiming to better capture semantic information and category distribution in the feature space ${\mathbf{x}_i}$.

\begin{figure*}[t]
  \centering
  \begin{tabular}{cc}
    \includegraphics[width=0.98\textwidth]{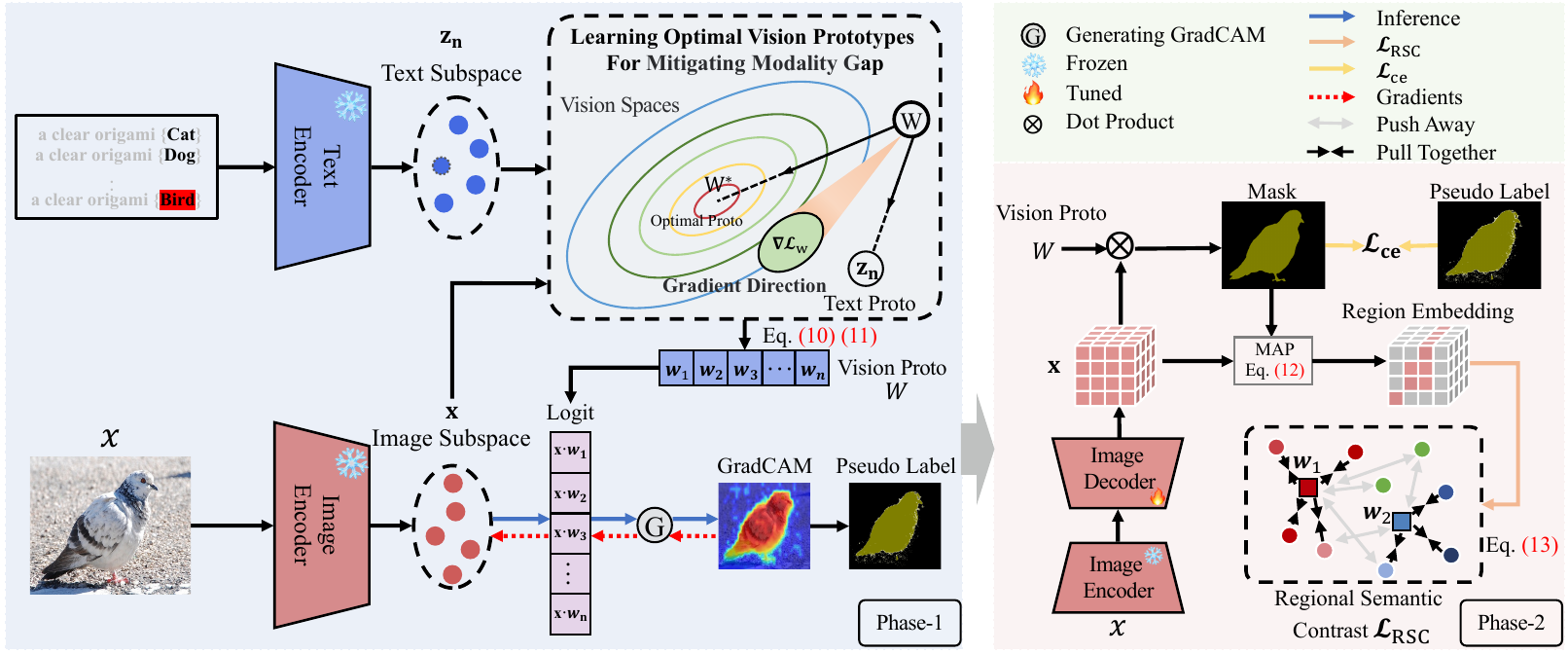}
  \end{tabular}
  \caption{Overview of the proposed Weakly-supervised Vision Prototype Learning (VPL), which consists of two main components: (1) Learning the vision prototype $W$ and (2) Regional semantic contrast (RSC). In phase (1), vision prototypes can be efficiently obtained by solving a convex optimization problem using gradient descent in Eq.~\eqref{17}. This ensures that it can align with vision data better to mitigate the impacts of the modality gap. The vision prototypes will then replace text prototypes to locate the target object and generate initial GradCAMs. In phase (2), we refine these GradCAMs to form pseudo-labels for supervising the decoder while CLIP encoders are frozen. Then, we obtain the masks from the vision prototypes and align them with specific region embeddings.}
  \label{method}
  
\end{figure*}
\noindent\textbf{We raise a question: Is it possible to find the optimal vision prototypes within the text space?}
\begin{hypothesis}
We decompose the class prototype as:
\begin{equation}\small
\begin{aligned}
\mathbf{z}_n &= \sqrt{\epsilon} \mathbf{z}_n^x + \sqrt{1-\epsilon} \mathbf{z}_n^{\perp}, \\
\text{s.t.} \quad \left\|\mathbf{z}_n^x\right\| = \left\|\mathbf{z}_n^{\perp}\right\| &=1, \quad \mathbf{z}_n^{x \top} \mathbf{z}_n^{\perp}=0, \quad 0 \leq \epsilon \leq 1,
\end{aligned}
\end{equation}
where $\mathbf{z}_{n}^{x}$ is from the vision space spanned by $\{\mathbf{x}_i\}$, and $\mathbf{z}_{n}^{\perp}$ represents the component from the orthogonal subspace with the condition that $\mathbf{z}_{n}^{x}{}^\top \mathbf{z}_{n}^{\perp} = 0$, and they have unit norms. We note that $\lim_{\epsilon \to 1} \mathbf{z}_n = \mathbf{z}_n^x$. In this case, $\mathbf{z}_n$ retains more vision information.
\label{hypothesis1}
\end{hypothesis}
\noindent\textbf{Remark:} The magnitude of $\epsilon$ determines the overlap between two components. Our hypothesis suggests that with this decomposition, if the vision space is covered by the text space, such that the class prototype $\mathbf{z}_n$ contains all information from both modalities, then it is possible to recover optimal segmentation performance. The details are as follows.

% \noindent\textbf{Remark:} The value of $\epsilon$ controls the overlap between the two components. Our hypothesis is that if the vision space is encompassed by the text space, allowing the class prototype $\mathbf{z}_n$ to integrate information from both modalities, optimal segmentation performance can be achieved. Details are provided below.

\begin{proposition} Define $p_{i, k}^n$ and $p_{i, k}^{n \prime}$ respectively denote the predicted probabilities for each pixel $k$ obtained by the vision prototypes $W^*$ and text prototypes $Z$. Then, we have:
\begin{equation}\small
p_{i, k}^{n \prime}=\frac{\exp \left(\mathbf{x}_{i,k}^{\top} \mathbf{z}_n / \tau_T\right)}{\sum_n \exp \left(\mathbf{x}_{i,k}^{\top} \mathbf{z}_n / \tau_T\right)}=\frac{\exp \left(\sqrt{\epsilon} \mathbf{x}_{i,k}^{\top} \mathbf{w}_n^* / \tau_T\right)}{\sum_n \exp \left(\sqrt{\epsilon} \mathbf{x}_{i,k}^{\top} \mathbf{w}_n^* / \tau_T\right)},
\end{equation}
where $\tau_T$ represents the temperature in CLIP, while $\tau_I$ is the temperature for learning vision prototypes. If $\mathbf{z}_n^{x}=\mathbf{w}_n^{*}$, then $p_{i, k}^{n \prime}=p_{i, k}^n$ and $\tau_T=\sqrt{\epsilon}\tau_I$. See proof in Appendix B.3. \qed
\label{space}
\end{proposition}
\noindent \textbf{Remark:} Proposition~\ref{space} suggests that if the text space encompasses the entire vision space, a larger temperature $\tau_T=\sqrt{\epsilon}\tau_I$ can help identify the optimal vision prototypes within the text space by smoothing the probability distribution and narrowing the modality gap.

Despite significant progress in integrating natural language and vision information, the text and vision spaces learned by CLIP are distinct, with a clear margin, as illustrated in Fig.~\ref{intro} (a). It is difficult for the text space to cover the vision space. The assumptions (Hypothesis~\ref{hypothesis1} and Proposition~\ref{space}) suggesting that the text space can recover the optimal vision prediction are difficult to hold. This is due to the lack of sufficient detail in textual data to precisely describe the information in visual data (\textit{e.g.,} colors and textures). Thus, the capability of text prototypes to provide precise guidance for pixel-level classification is limited. The lower bound for the modality gap is quantified in the following theorem.

\begin{theorem}[The lower-bound of modality gap]
The image and text embeddings of CLIP are always on the unit sphere, with the Euclidean distance $\|Z-W^*\|^2$, being defined as:
\begin{equation}\small
\begin{aligned}\Delta_{gap} = \left\|Z-W^*\right\|_F^2 \geq \boldsymbol{c}_{\perp}+  \boldsymbol{\epsilon}, 
\end{aligned}
\end{equation}
where $\boldsymbol{c}_{\perp}$ denote a constant vector, which is orthogonal to the embedding span of $Z$ and $W^*$. And $\boldsymbol{\epsilon}$ representing the alignment noise. See proof details in Appendix B.4. \qed
\label{theorem1}
\end{theorem}

\noindent\textbf{Remark:} Theorem \ref{theorem1} emphasizes the modality gap between the text prototype and the optimal vision prototype and consists of two components. $\boldsymbol{c}_{\perp}$ represents the distance to the irrelevant text space, \textit{i.e.,} redundant text dimensions. This implies that within the text space, there exist dimensions or information that are irrelevant to the vision modality. $\boldsymbol{\epsilon}$ represents the approximation loss caused by the low-rank overlap between the text and vision spaces, which indicates that there may be a certain degree of overlap between the text and vision spaces. They share some information but cannot match each other completely, indicating the presence of partial semantic information missing in the text prototype. Thus, minimizing the distance between $Z$ and $W$ is difficult due to the inherent modality gap, which confirms our empirical observation.

To accurately capture vision semantic features, we obtain vision prototypes in the following Theorem~\ref{theorem2}. Since without precise pixel-level annotations, object masks in Eq.~\eqref{33} can be generated by text prototypes using GradCAM. Our objective function is constructed by measuring the KL-divergence between the distributions of text prototypes and vision prototypes as:
\begin{equation}\small
\min _W \mathcal{L}\left(\mathcal{P}^{\prime}, W\right)=\sum_i {\mathrm{KL}}\left(\mathcal{P}_i^{n\prime} \| \mathcal{P}_i^n\right), \label{13}
\end{equation}
where $\mathcal{P}_i^{n\prime}$ and $\mathcal{P}_i^{n}$ respectively represent the distributions estimated by the text prototypes $Z$ and the learnable vision prototypes $W$. This constraint aligns the model within a common representation space, reducing the modality gap.

Furthermore, we optimize the KL divergence in two ways: one is class-level regularization, where given anchor class $n$, the distribution across all categories can be computed. The other is instance-level distribution regularization, where the anchor class $n$ is replaced with anchor $\mathbf{x}_i$.
\begin{theorem}\label{theorem42}
Assuming the norm of prototypes is bounded by $\eta$, \textit{i.e.,} $\forall n,\|\mathbf{w}_n\|_2 \leq \eta$, the distribution defined by the anchor class is an approximation of the distribution defined by the anchor example as:
\begin{equation}\small
\forall n, \quad \frac{1}{\exp \left(2 \kappa\right)} \mathcal{P}_{m_{i,k}} \leq \mathcal{P}_{i, j} \leq \exp \left(2 \kappa\right) \mathcal{P}_{m_{i,k}},
\end{equation} 
where $\kappa = 2 \eta\left\|\mathbf{x}_i-\mathbf{w}_{m_{i,k}}\right\|_2$. When the intra-class distribution is compact, \textit{i.e.,} $\|\mathbf{x}_i-\mathbf{w}_{m_{i,k}}\|_2 \rightarrow 0$, this approximation becomes tight. See proof details in Appendix B.5.
\qed
\label{theorem2}
\end{theorem}
\noindent \textbf{Remark:} Theorem~\ref{theorem42} demonstrates that instance-level regularization, as opposed to class-level regularization, can help better capture variations in real data.

\begin{theorem}[Learning Vision Prototypes] Assume $\mathcal{L}\left(\mathcal{P}^{\prime}, W\right)$ is a $\mu$-strongly convex function in $W$, we have:
\begin{equation}\small
\begin{aligned}
W^{\prime *} & =\underset{W}{\arg \min } \ \mathcal{L}\left(\mathcal{P}^{\prime}, W\right); \quad W^*=\underset{W}{\arg \min }\ \mathcal{L}(m, W),
\end{aligned}
\label{15}
\end{equation} 
where $m_i$ is the GradCAM as the pseudo mask distribution for $\mathbf{x}_i$, and is generated through text prototypes $Z$. We compute the Eq.~\eqref{13} as $-\sum_{i, n,k} \mathcal{P}_{i, k}^{n\prime} \log (\mathcal{P}_{i, k}^n)$, and we compute the gradient direction to adjust $W^{\prime *}$ by the standard gradient descent. Subsequently, we have:
\begin{equation}\small
\begin{aligned}
\nabla_{W^{\prime *}} \mathcal{L}_{\left(W\right)} & \leq \frac{2}{\mu}\left\langle \mathcal{P}^{\prime}-Y, \log \left(\mathcal{P}_{W^{\prime *}}\right)-\log \left(\mathcal{P}_{W^*}\right)\right\rangle, 
\end{aligned} \label{16}
\end{equation}
where $\nabla_{W^{\prime *}} \mathcal{L}_{\left(W\right)}$ denote $\|W^{\prime*}-W^* \|^2$. Then, we update the $W^{\prime *}$ to minimize the loss function:
\begin{equation}\small
W^{\prime *} \leftarrow W^{\prime *}-\alpha \nabla_{W^{\prime *}} \mathcal{L}_{\left(W\right)} \label{17}
\end{equation}
where $\alpha$ is the learning rate, we iterate $T_w$ times to gradually optimize the parameters. Update $W^{\prime *}$ in the negative gradient direction to obtain a local optimal solution for the vision prototype.
See proof details in Appendix B.6. \qed
\label{theorem2}
\end{theorem}

\begin{algorithm}[tb]
    \caption{Learning the Optimal Vision Prototypes}
    \label{alg:algorithm}
    \textbf{Require}: Input image set $\left\{x_i\right\}$, class names $\left\{z_n\right\}$, paired image text encoder $E_v, E_t$, vision prototype $W^{\prime*}$, loss function $\mathcal{L}$, iterations $T_w$, temperature $\tau_T,\tau_I$, learning rate $\alpha$
    \begin{algorithmic}[1]
        \STATE $\mathbf{x}_i$$\leftarrow$$E_v\left(x_i\right), \mathbf{z}_n$$\leftarrow$$E_t\left(z_n\right)$  \hfill  $\triangleright$ See Eq.~\eqref{extracted}
        \STATE $W^{\prime*}$ $\leftarrow$ $\min_W \mathcal{L}\left(P^{\prime}, W\right)$ \hfill $\triangleright$ See Eq.~\eqref{15}
        \STATE Calculate gradient $\text{Grad}=\nabla_{W^{\prime *}} \mathcal{L}_{\left(W\right)}$ \hfill $\triangleright$ See Eq.~\eqref{16}
        \STATE $W^{\prime *} \leftarrow W^{\prime *}-\alpha \nabla_{W^{\prime *}} \mathcal{L}_{\left(W\right)}$ \hfill $\triangleright$ See Eq.~\eqref{17}
        \STATE Calculate GradCAMs $M_n$ using $W^{\prime *}$ \hfill $\triangleright$ GradCAM calculation (See Eq.~\eqref{extracted})
        \STATE \textbf{return} $W^{\prime *}$, $M_n$
    \end{algorithmic}
\end{algorithm}

\noindent \textbf{Remark:} The core of our method is using KL-divergence and strong convexity to efficiently approach the optimal solution by gradient descent. By Proposition~\ref{space}, the temperature $\tau_I$ during the learning of vision prototypes should be larger than $\tau_T$ in CLIP, and has been confirmed in our ablation study. Our experiments show that our vision prototype significantly improves the accuracy of target localization compared to the use of text prototypes with single prompt or ensemble prompts. As illustrated in Fig.~\ref{intro} (b), the GradCAM generated by the vision prototypes focuses more on the target object rather than the background or irrelevant regions. The detailed algorithmic is given in Alg.~\ref{alg:algorithm}.

\textbf{Regional Semantic Contrast Learning.} To learn the alignment between pixel-level features and prototypes, we supervise the mask decoder using post-processing GradCAM. However, relying solely on this loss is insufficient to fully exploit object regions as the noise in pseudo labels leads to errors in segmentation. Therefore, we propose a regional semantic contrast module to effectively contrast categorical object regions and vision prototypes by suppressing negative masks obtained from unrelated texts (\textit{i.e.,} texts of negative pairs). Specifically, for the pixel-level dense embeddings $\mathbf{V} \in \mathbb{R}^{B\times D\times H \times S}$ from the decoder and the vision prototypes $W \in \mathbb{R}^{D \times N}$, we compute the feature-level localized image embeddings $P_{i, n} \in \mathbb{R}^D$ by masked average pooling (MAP) as follow: 
\begin{equation}\small
P_{i, n}=\frac{\sum_{h, w} \mathbf{A}_{i, n, h, w} \cdot \mathbf{V}_{i, :, h, w}}{\sum_{h, w} \mathbf{A}_{i, n, h, w}},
\end{equation}
where $B$ denotes a batch size. $\mathbf{A}_{i,n} = W_n^{\top} \mathbf{V}_i$ calculates the masks in the batch. Subsequently, we calculate the cosine similarity between vision prototypes and feature-level localized image embeddings as $S_{i, n}=P_{i, n}^{\top} \mathbf{W}_n$. RSC Loss is defined as follows:
\begin{equation}\small
\begin{aligned}
\mathcal{L}_{\mathrm{RSC}}= & -\frac{1}{2 B} \sum_i^B \log \frac{\exp \left(S_{i, i} / \tau\right)}{\sum_j^B \exp \left(S_{i, j} / \tau\right)} \\
& -\frac{1}{2 B} \sum_i^B \log \frac{\exp \left(S_{i, i} / \tau\right)}{\sum_j^B \exp \left(S_{j, i} / \tau\right)}.
\end{aligned}
\end{equation}
\textbf{Final Loss.} Our final loss function is defined by:
\begin{equation}\small
\mathcal{L}={\lambda_{\mathrm{ce}} \mathcal{L}_{\mathrm{ce}}(M,\mathbf{A})} + {\lambda_{\mathrm{RSC}} \mathcal{L}_{\mathrm{RSC}}}, 
\label{coefficients}
\end{equation}
where $\mathcal{L}_{\mathrm{ce}}$ is the cross-entropy loss and $\mathcal{L}_{\mathrm{RSC}}$ is the region-level RSC loss. $M$ denotes the mask obtained by post-processing the GradCAM. $\lambda_{\mathrm{ce}}$ and $\lambda_{\mathrm{RSC}}$ are coefficients.

\section{Experiments}

\label{sec:Experiments}
\subsection{Datasets and Implementation Details}

\textbf{Datasets and Evaluation Metrics.} Experiments are conducted on two benchmarks: PASCAL VOC 2012 \cite{everingham2010pascal} with 21 classes and MS COCO 2014 \cite{lin2014microsoft} with 81 classes. For PASCAL VOC 2012, following \cite{wang2020self,lee2021anti,li2022expansion}, we use the augmented SBD \cite{hariharan2011semantic} with 10,582 annotated images. We evaluate VPL in terms of i) the quality of generated pseudo masks on VOC 2012 \texttt{train}, and ii)semantic segmentation on VOC 2012 \texttt{val/test} and COCO 2014 \texttt{val}. Mean intersection over union (mIoU) \cite{long2015fully} is used as the metric in both cases. VOC 2012 \texttt{test} scores are obtained from the official evaluation server.

\noindent\textbf{Implementation Details.} In our experiments, we adopt the CLIP pre-trained model ViT-B-16~\cite{radford2021learning}. The feature map used to generate CAM is the one before the last self-attention layer in ViT. We replace the class token with the average of remaining tokens to compute final logits. For the VOC 2012 dataset, phase 1 of our framework is learned by standard projected gradient descent, where the learning rate is 10 and the number of iterations is 3,000. $\tau_T$ is 0.01 in CLIP and $\tau_I$ for learning the vision prototypes is set to 0.03 in Section~\ref{3.3}. CAMs are refined into pseudo masks using denseCRF~\cite{krahenbuhl2011efficient}. In phase 2, the batch size is set as 6 and the maximum iteration is set as 30,000. SGD optimizer is adopted to train only our decoder with a momentum of 0.9 and a weight decay of 1e-4. All experiments are conducted on 8 Nvidia 4090 GPUs. The loss coefficients $\lambda_{\mathrm{ce}}$ and $\lambda_{\mathrm{RSC}}$ are both set as 1 in Eq.~\ref{coefficients}.

\begin{table}[t]
\centering
\resizebox{\columnwidth}{!}{
\begin{tabular}{lcc}
\toprule[1pt]
\textbf{Method} & \textbf{Seed}  & \textbf{Pseudo-Mask} \\ 
\toprule[1pt]
$\textnormal{IRN}$
\cite{ahn2019weakly} & {48.8}  & {66.5}\\
$\textnormal{ESOL}$~\cite{li2022expansion} & 53.6  & 68.7 \\ 
$\textnormal{MCTformer}$~\cite{xu2022multi} & 61.7 & 69.1\\
$\textnormal{LPCAM}$~\cite{chen2023extracting} & 65.3 &  72.7 \\ 
$\textnormal{ACR}$~\cite{kweon2023weakly} & 60.3 & 72.3 \\ 
$\textnormal{D2CAM}$~\cite{wang2023treating} & 58.0 & 71.4 \\
$\textnormal{Mat-Label}$~\cite{wang2023treating} & 62.3 & 72.9\\ 
$\textnormal{FPR}$~\cite{chen2023fpr} & 67.7 & 72.8 \\ 
$\textnormal{USAGE}$~\cite{peng2023usage} & 71.9 & 72.8 \\
$\textnormal{SFC}$~\cite{zhao2024sfc} & 64.7 & 73.7 \\
$\textnormal{SeCo}$~\cite{yang2024separate} & 74.8 & 76.5 \\ 
$\textnormal{SFC}$~\cite{zhao2024psdpm} & 64.7 & 73.7 \\
$\textnormal{DuPL}$~\cite{wu2024dupl} & 73.5 & 75.1 \\
$\textnormal{PSDPM}$~\cite{zhao2024psdpm} & 68.5 & 77.3 \\ \midrule
$\textnormal{CLIP-ES}$~\cite{lin2023clip} &70.8  & 75.0 \\ \rowcolor{gray!30}
\textbf{+VPL (Ours)}  &\textbf{76.3}\textcolor{blue!60}{$\uparrow$5.5}  & \textbf{78.4}\textcolor{blue!60}{$\uparrow$3.4} \\
\midrule
$\textnormal{CLIP-CPAL}$~\cite{tang2024hunting} & 71.9 & 75.8 \\ \rowcolor{gray!30}
\textbf{+VPL (Ours)}  &\textbf{77.8}\textcolor{blue!60}{$\uparrow$5.9}   & \textbf{80.1}\textcolor{blue!60}{$\uparrow$4.3} \\
\toprule[1pt]
\end{tabular}}
\caption{Comparisons between our method and the other WSSS methods. Evaluate mIoU (\%) on the PASCAL VOC 2012 \texttt{train} set at levels: CAM and Pseudo-Mask. The best results are in bold.}
\label{labelVOC}
\end{table}

\begin{figure}[!h]
\centering
\resizebox{\columnwidth}{!}{  
\begin{tabular}{cccc}  
\includegraphics[width=0.35\textwidth]{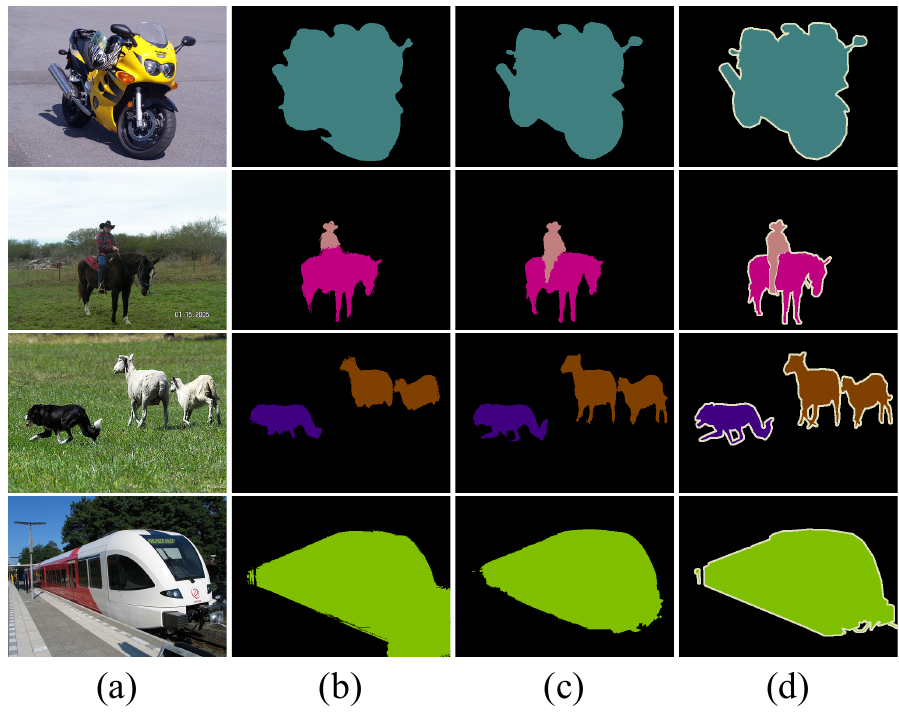}
\end{tabular}}
\caption{\small{Qualitative results on Pascal VOC 2012 \textit{val} set. (a) Input images. (b) Results from CLIP-ES. (c) Results by our CLIP-ES+VPL. (d) Ground truth. Our method produces more accurate responses and as a plug-and-play method.}}
\label{quali}
\end{figure}

\begin{table}[t]
\centering
\resizebox{\columnwidth}{!}{
\begin{tabular}{lcccl}
\toprule[1pt]
\multirow{2}{*}{{\textbf{Methods}}} & \multicolumn{2}{c}{{\textbf{VOC}}} & \multicolumn{1}{c}{{\textbf{COCO}}} \\ \cmidrule(r){2-4} 
 & \textbf{Val} & \textbf{Test} & \textbf{Val} \\   
\toprule[1pt]
\multicolumn{3}{l}{\textbf{Image-level supervision.}} \\
$\textnormal{IRN}$~\cite{ahn2019weakly}   & 63.5 & 64.8 & 41.4 \\
$\textnormal{OCR-SEAM}$~\cite{cheng2023out}   & 67.8 & 68.4 & 33.2 \\
$\textnormal{ACR}$~\cite{sun2023all}   &71.2 &70.9 & 45.0 \\
$\textnormal{ToCo}$~\cite{ru2023token}   & 71.1 & 72.2 & 42.3 \\ 
$\textnormal{LPCAM}$~\cite{chen2023extracting}   & 70.1 & 70.4 & 45.5 \\ 
$\textnormal{BECO}$~\cite{rong2023boundary}  & 73.7 & 73.5 & 45.1 \\
$\textnormal{DuPL}$~\cite{wu2024masked}  & 73.3 & 72.8 & 44.6 \\ \midrule
\multicolumn{3}{l}{\textbf{Image-level supervision + Language supervision.}} \\
$\textnormal{CLIMS}$~\cite{xie2022clims}    & 70.4 & 70.0 & - \\ 
$\textnormal{MMCST}$~\cite{xu2023learning}    & 72.2 & 72.2 & 45.9 \\
$\textnormal{WeCLIP}$~\cite{zhang2024frozen}    & 76.4 & 77.2 & 47.1 \\ \midrule
$\textnormal{CLIP-ES}$~\cite{lin2023clip}  & 73.8 & 73.9 & 45.4 \\ \rowcolor{gray!30}
\textbf{+VPL (Ours)} & \textbf{78.5}\textcolor{blue!60}{$\uparrow$4.7} & \textbf{77.8}\textcolor{blue!60}{$\uparrow$3.9} &   \textbf{49.2}\textcolor{blue!60}{$\uparrow$3.8} \\ \midrule
$\textnormal{CLIP-CPAL}$~\cite{tang2024hunting} & 74.5 & 74.7 & 46.8 \\ \rowcolor{gray!30}
\textbf{+VPL (Ours)} & \textbf{79.3}\textcolor{blue!60}{$\uparrow$4.8} & \textbf{79.0}\textcolor{blue!60}{$\uparrow$4.3} & \textbf{49.8}\textcolor{blue!60}{$\uparrow$3.0} \\
\toprule[1pt]
\end{tabular}}
\caption{\small{Evaluation the mIoU results (\%) on PASCAL VOC \texttt{val} and \texttt{test} sets, and COCO \texttt{val} set. The best results for each dataset are in bold.}} 
\label{miou_results}
\end{table}

\subsection{Comparisons With State-of-the-Art Methods}

\textbf{Improved Localization Maps and Segmentation Results:} The proposed CLIP-CPAL does not modify the architecture of the CLIP network. It simply integrates VPL into the CLIP-ES and CLIP-CPAL methods. The results on VOC 2012 show improved localization maps, as shown in Table~\ref{labelVOC}. For instance, incorporating VPL into CLIP-ES improves seed performance by 5.5\% and pseudo mask performance by 2.5\%. When VPL is added to the CLIP-CPAL model, there is a 5.9\% gain in seed performance. Table~\ref{miou_results} presents the performance of the semantic segmentation model trained with pseudo-labels generated by our method, compared with related works. Our CLIP-ES+VPL achieves sate-of-the-art (SOTA) results on VOC, with an mIoU of 78.5\% on the val set and 77.8\% on the test set. On the more challenging COCO dataset, our CLIP-ES+VPL surpasses the SOTA results of CLIP-ES and all related works. For CLIP-CPAL, VPL enhances performance (+3.0\% mIoU on the COCO validation set). The superior results on both datasets validate the effectiveness of our VPL in accurately capturing semantic features and object structures.

\noindent\textbf{Qualitative Results.} We present the qualitative results on the Pascal VOC dataset in Fig.~\ref{quali}. By observation, CLIP-ES using text prototypes poorly captures the complete outline of the object as shown in the $2^{\text{nd}}$ rows in Fig.~\ref{quali}. Also, in certain contexts, it has difficulty distinguishing between co-occurring categories, as shown in the $3^{\text{rd}}$ row, such as \texttt{train} and \texttt{railroad}). Equipped with VPL, the model predicted more accurately on shapes and classes.

\subsection{Ablation Study}\label{Ablation}

We conduct ablation studies on the PASCAL VOC 2012 \texttt{val} set and use CLIP-ES as the Baseline.

\begin{table}[t]
\centering
\resizebox{\columnwidth}{!}{
\begin{tabular}{c|c|c|c|c|c|c}
\toprule[1pt] & \textbf{Text Pro} & \textbf{Vision Pro} & \bm{$\mathcal{L}_{\mathrm{CE}}$} & \textbf{RSC\bm{$(\mathcal{L}_{\mathrm{RSC}})$}} & \textbf{CRF} & \textbf{mIoU(\%)} \\ \midrule\midrule
\text{I} & \CheckmarkBold & & & & & 63.3  \\
\text{II} & & \CheckmarkBold & & & & 68.2 \\
\text{III} &  & \CheckmarkBold &  & & \CheckmarkBold & 71.3 \\
\text{IV} &  & \CheckmarkBold  & \CheckmarkBold & & & 75.5 \\
\text{V} & &  \CheckmarkBold & \CheckmarkBold & \CheckmarkBold & & 77.2 \\
\text{VI} & &  \CheckmarkBold & \CheckmarkBold & \CheckmarkBold & \CheckmarkBold & \textbf{78.5} \\
\toprule[1pt]
\end{tabular}}
\caption{\small{Ablation study on main components of the proposed framework. The mIoU values are evaluated on the PASCAL VOC 2012 \texttt{val} set. In phase 1, Text: Baseline only uses text prototypes. Vision: Vision prototypes for dense localization. CRF: Adoption for post-processing. In phase 2, $\mathcal{L}_{\mathrm{CE}}$: Supervise the network with pseudo-mask. RSC$(\mathcal{L}_{\mathrm{RSC}})$: Align region embedding with the vision prototypes.}} \label{abl}
\end{table}

\begin{table}[t]
\centering
\resizebox{\columnwidth}{!}{
\begin{tabular}{lcc}
\toprule[1pt]
\textbf{Methods} & \textbf{mIoU(\%)} & \textbf{Gap}  \\ \midrule \midrule
$\text{Baseline}_{\text{s}}$ & 61.7 & 0.81 \\
$\text{Baseline}_{\text{e}}$  & 63.3 & 0.76 \\ \toprule[1pt]
Modify gap~\cite{liang2022mind} +s & 62.0 & 0.77 \\ 
Modify gap~\cite{liang2022mind} +e & 63.5 & 0.71 \\ \toprule[1pt]
$\text{VPL}_{\text{s}}$     & 66.8  & 0.64 \\ \rowcolor{gray!30}
$\text{VPL}_{\text{e}}$     & \textbf{68.2} & \textbf{0.51} \\ \toprule[1pt]
\end{tabular}}
\caption{\small{Comparison of mIou(\%) with different text prompts. ``s'' denotes a single prompt, while ``e'' indicates ensemble prompts. ``Modify gap" refers to reducing the differences between modalities by adjusting their vector representations~\cite{liang2022mind}.}}
\label{Prompt}
\end{table}

\noindent\textbf{Effectiveness of each component.} Table~\ref{abl} presents ablation experiments demonstrating the effectiveness of VPL in two phases. In Experiment I, the model is supervised exclusively using text prototypes, which serve as the primary baseline. In Experiment II, learning vision prototypes to generate GradCAMs  enhances performance, with an increase of +4.9\%. This indicates that vision prototypes capture fine-grained semantic details better than text prototypes. Subsequently, using CRF to generate pseudo masks boosts performance to 71.3\% (Experiment III). In phase 2, we introduce cross-entropy loss to supervise the segmentation network (Experiment IV). Specifically, we freeze the CLIP encoder and only train the decoder, resulting in a further increase of +4.2\% in performance. In Experiment V, when regional semantic contrast is used as complementary supervision, performance is further improved by +1.7\%, indicating its importance in our framework. This module enables the model to focus on the target region, enhancing the matching degree between prototypes and region embeddings.

\begin{table}[t]
\centering
% \resizebox{\columnwidth}{!}{
\begin{tabular}{ccc}
\toprule[1pt]
$\tau_I$ & \textbf{mIoU(\%)} & \textbf{Gap} \\
\toprule[1pt]
Baseline & 63.3 & 0.76 \\
0.01     & 66.9 & 0.64 \\
0.02     & 67.3 & 0.57 \\ \rowcolor{gray!30}
0.03     & \textcolor{red}{68.2} & 0.51 \\
0.04     & 67.8 & 0.49 \\
0.05     & 67.2 & \textcolor{red}{0.47} \\
\toprule[1pt]
\end{tabular}
\caption{Impact of temperature $\tau_I$ on PASCAL VOC.}
\label{table5}
\end{table}

\begin{table}[t]
\centering
\resizebox{\columnwidth}{!}{
\begin{tabular}{ccccccc}
\toprule[1pt]
\bm{$\Phi$} & 0.2 & 0.3 & 0.4 & 0.5 & 0.6 & 0.7 \\
\toprule[1pt]
\textbf{mIoU(\%)} & 67.2 & 67.7 & \cellcolor{gray!30}\textcolor{red}{68.2} & 67.9 & 67.3 & 67.9 \\
\toprule[1pt]
\end{tabular}}
\caption{Impact of Threshold $\Phi$ on PASCAL VOC.}
\label{table6}
\end{table}

\begin{table}[!h]
\centering
\resizebox{\columnwidth}{!}{
\begin{tabular}{ccccccc}
\toprule[1pt]
\bm{$T_w$} & 1000 & 1500 & 2000 & 2500 & 3000 & 3500 \\
\toprule[1pt]
\textbf{mIoU(\%)} & 65.9 & 66.6 & 67.5 & 67.8 & \cellcolor{gray!30}\textcolor{red}{68.2} & 68.0 \\
\toprule[1pt]
\end{tabular}}
\caption{Impact of Iteration $T_w$ on PASCAL VOC.}
\label{table7}
\end{table}

\noindent\textbf{Impact of temperature.} To investigate the impact of the temperature $\tau_I$ in learning vision prototypes, Table~\ref{table5} shows the mIoU and Gap results at different temperatures on the PASCAL VOC 2012 \texttt{val} set. The Gap measures the difference between various prototypes and features, with a lower gap indicating higher similarity. “Baseline” refers to the generation of CAMs using text prototypes. At $\tau_I=0.03$, the vision prototypes already show an improvement by +4.9\% compared to text prototypes. As $\tau_I$ increases, the modality gap decreases, enhancing the localization results. This also confirms Proposition~\ref{space}, which suggests $\tau_I$ should be larger than $\tau_T=0.01$ in CLIP. This indicates that vision and text spaces share common information. By tuning $\tau_I$, we can better utilize this shared information. Thus, we fix $\tau_I=0.03$ as the optimal value to improve performance.

\begin{figure}[!h]
\centering
\includegraphics[width=0.45\textwidth]{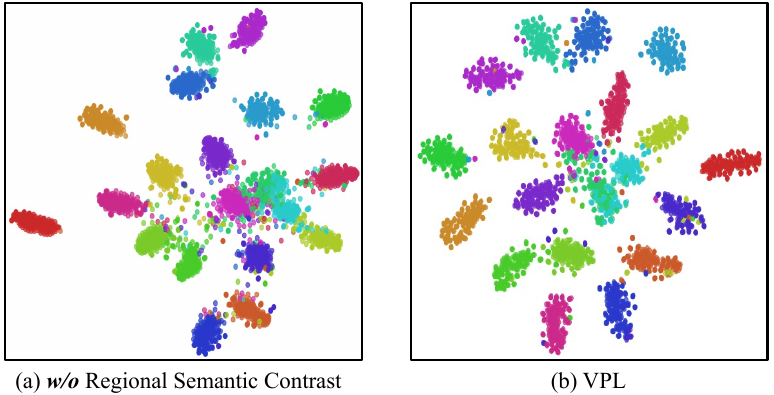}
\caption{\small{Feature embedding visualizations of (a) our framework without RSC, and (b) our framework on the Pascal VOC 2012 \texttt{val} set using t-SNE~\protect\cite{van2008visualizing}.}}
\label{tsnet}
\end{figure}

\noindent\textbf{Prompt Selection.}  In Table~\ref{Prompt}, we compare two prompt strategies: using a single prompt like \textit{“a photo of a [CLS]”} and using an ensemble of multiple prompts such as \textit{“a clean origami [CLS].”} The placeholder \textit{[CLS]} represents a category label and its synonyms. Our VPL with the ensemble prompt strategy improves performance over the single prompt by +1.4\%. Furthermore, our method outperforms both prompts on text prototypes (``Baseline"), demonstrating the effectiveness of vision prototypes.

\noindent\textbf{Analysis of Hyper-parameters.} We conduct a hyperparameter analysis for different values, such as the threshold $\Phi$ for generating 0-1 seed masks. In Table~\ref{table7} (b), the threshold $\Phi$ is varied from 0.2 to 0.7 with an interval of 0.1. We find that the optimal choice $\Phi$ is 0.4. Moreover, we also investigate the number of iterations $T_w$ for gradient descent to learn the vision prototype and find that the highest accuracy of 68.2\% is achieved with 3,000 iterations in Table~\ref{table7} (c).

\noindent\textbf{Effectiveness of RSC.} In Table~\ref{abl}, we present the performance improvement results achieved through the regional semantic contrast module. Moreover, we conduct a visual comparison using t-SNE~\cite{van2008visualizing} in Fig.~\ref{tsnet}. The results indicate that after aligning regional features and prototypes, the model is capable of generating more compact clusters with increased inter-cluster separability. 

\section{Conclusion} 
In this work, we focus on improving the capability of CLIP in weakly supervised semantic segmentation. Our theoretical analysis indicates that the modality gap between text and vision spaces obtained in CLIP is inherent, which can impair the performance. To alleviate this, we propose a Vision Prototype Learning (VPL) framework, which learns more representative prototypes in the vision space and generates high-quality segmentation masks. Extensive experiments have demonstrated that our framework achieves state-of-the-art performance and has the potential to segment new classes.

\section{Acknowledgments}
This research was supported by an Australian Government Research Training Program (RTP) Scholarship awarded to Zhongxing Xu.

\bibliography{aaai25}
\end{document}